\documentclass[conference]{IEEEtran}
\IEEEoverridecommandlockouts
\usepackage{cite}
\usepackage{amsmath,amssymb,amsfonts}
\usepackage{algorithm}
\usepackage{algorithmic}
\usepackage{graphicx}
\usepackage{textcomp}
\usepackage{url}
\usepackage{xcolor}
\usepackage{caption}
\usepackage{subcaption}
\usepackage{threeparttable, booktabs}
\usepackage{tabularx}
\usepackage{tabularx} 
\usepackage{array}    
\usepackage{booktabs} 
\usepackage[a4paper, total={184mm,239mm}]{geometry}
\def\BibTeX{{\rm B\kern-.05em{\sc i\kern-.025em b}\kern-.08em
    T\kern-.1667em\lower.7ex\hbox{E}\kern-.125emX}}
\usepackage{hyperref}
\hypersetup{
    colorlinks=true,    
    urlcolor=blue,      
    linkcolor=blue,     
    citecolor=blue      
}
\definecolor{deepblue}{rgb}{1.0, 0.341, 0.2}
\definecolor{color2}{rgb}{0.290, 0.561, 0.882}

\begin{document}


\title{
One Step Beyond: Feedthrough \& Placement-Aware Rectilinear Floorplanner
}




\author{\IEEEauthorblockN{Anomymous Authors}}
\author{\IEEEauthorblockN{
Zhexuan Xu$^1$, Jie Wang$^1$, Siyuan Xu$^2$, Zijie Geng$^1$, Mingxuan Yuan$^2$ and Feng Wu$^1$
}
  $^{1}$University of Science and Technology of China\\
  $^{2}$Noah’s Ark Lab, Huawei\\
}

\maketitle

\begin{abstract}
Floorplanning determines the shapes and locations of modules on a chip canvas and plays a critical role in optimizing the chip's Power, Performance, and Area (PPA) metrics. However, existing floorplanning approaches often fail to integrate with subsequent physical design stages, leading to suboptimal in-module component placement and excessive inter-module feedthrough. To tackle this challenge, we propose Flora, a three-stage feedthrough and placement aware rectilinear floorplanner. In the first stage, Flora employs wiremask and position mask techniques to achieve coarse-grained optimization of HPWL and feedthrough. In the second stage, under the constraint of a fixed outline, Flora achieves a zero-whitespace layout by locally resizing module shapes, thereby performing fine-grained optimization of feedthrough and improving component placement. In the third stage, Flora utilizes a fast tree search-based method to efficiently place components—including macros and standard cells—within each module, subsequently adjusting module boundaries based on the placement results to enable cross-stage optimization. Experimental results show that Flora outperforms recent state-of-the-art floorplanning approaches, achieving an average reduction of 6\% in HPWL, 5.16\% in FTpin, 29.15\% in FTmod, and a 14\% improvement in component placement performance.
\end{abstract}


\section{Introduction}

As the scale of modern Very Large Scale Integration (VLSI) circuits continues to grow, Electronic Design Automation (EDA) has become an indispensable tool for managing these complexities. Floorplanning, a foundational step in the EDA physical design flow, aims to determine the shapes and positions of circuit modules without overlap. These decisions directly impact the detailed placement stage—where components such as larger macros and smaller standard cells are positioned within modules—and the routing stage—where physical paths are established for modules and components across the layout. The overarching objective of these stages is to enhance the chip’s Power, Performance, and Area (PPA) metrics.

Existing floorplanning approaches share a significant limitation: they focus exclusively on the floorplanning stage, prioritizing half-perimeter wire length (HPWL) as the primary optimization metric, while overlooking the subsequent physical design process—particularly metrics related to feedthrough (a routing technique for long-distance connections) and component placement. If floorplanning considers only the placement of modules without accounting for subsequent stages, it can lead to inconsistencies in the EDA physical design flow, ultimately resulting in suboptimal design outcomes.

\begin{figure}[t]
    \centering
    \includegraphics[width=\linewidth]{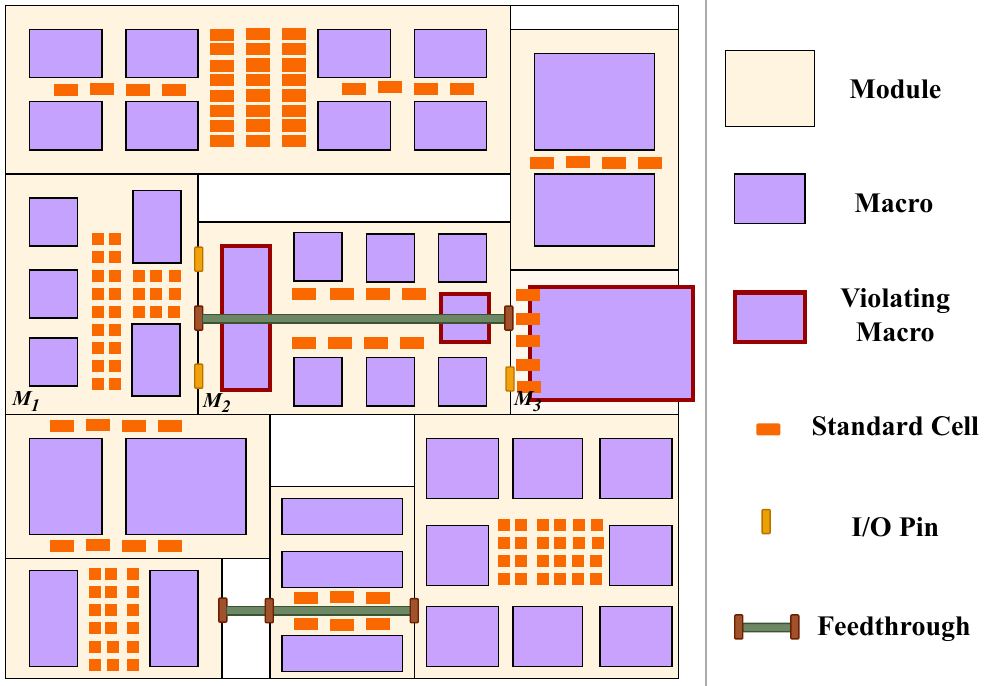}
    \caption{\textbf{Motivating example.} A simplified illustration of floorplanning and component placement involving feedthrough. In the current layout, a large macro cannot be placed inside module $M_3$. Moreover, the feedthrough between modules $M_1$ and $M_3$ compromises the internal integrity of module $M_2$.}
    \label{fig:motivation}
\end{figure}

Figure~\ref{fig:motivation} illustrates the motivation for our research. Many existing approaches perform detailed placement based on floorplanning layouts. On one hand, this can lead to designs that inadequately accommodate larger macros, as showed by the $M_3$ module. On the other hand, both the floorplanning and detailed placement stages rely on HPWL as the core optimization metric, which does not accurately reflect actual routing conditions. In practice, a routing net such as \(\mathcal{N} = \{M_1, M_3\}\) requires feedthrough to connect non-adjacent modules in the physical layout. For simplicity, we depict a straight feedthrough in this case. However, the use of feedthrough introduces challenges, including compromised module integrity, longer routing paths, and signal attenuation~\cite{feedthrough}. Thus, effective floorplanning must establish a robust foundation for subsequent detailed placement and routing stages by addressing both component placement and feedthrough, ultimately optimizing PPA metrics.

To address these challenges, we introduce Flora, a \underline{f}eedthrough \& p\underline{l}acement-aware rectilinear flo\underline{or}pl\underline{a}nner that accounts for in-module component placement and inter-module feedthrough to pre-optimize subsequent physical design stages. Flora operates in three stages. First, it employs a simulated annealing approach that uses a wiremask to coarsely optimize the HPWL and feedthrough, while a position mask ensures legality of the layout. 
Second, it performs fine-grained optimization by resizing modules and allocating whitespace effectively, thereby further reducing feedthrough and enhancing component placement, ultimately achieving a zero-whitespace layout. 
Third, it strategically places components within each module—including large macros and small standard cells—using a tree search method, and subsequently refines module boundaries based on placement results to optimize overall component placement. Through Flora, we aim to provide a solid foundation for the subsequent detailed placement and routing stages, facilitating improved performance in these processes.

Our contributions are summarized as follows:
\begin{enumerate}
    \item We address the limitations of single-stage optimization by incorporating the subsequent physical design process into our framework, optimizing and evaluating both feedthrough and component placement.
    \item We propose a three-stage rectilinear floorplanning optimization framework that coarsely adjusts module positions, finely resizes modules, and rapidly places components while refining boundaries, achieving joint optimization of HPWL, feedthrough, and component placement.
    \item Flora serves not only as an end-to-end floorplanning framework but also as a post-processing enhancement for existing floorplanning algorithms.
    \item Experimental results on widely used benchmarks demonstrate Flora’s effectiveness in optimizing multiple metrics, including HPWL, whitespace, feedthrough, and component placement.
\end{enumerate}

The remainder of this paper is organized as follows: Section \ref{sec:previous work} reviews previous work; Section \ref{sec:preliminary} provides essential background on floorplanning; Section \ref{sec:propose method} details our proposed methodology; Section \ref{sec:experiment} presents the experimental results; and Section \ref{sec:conclusion} concludes the paper.

\section{Previous Work}\label{sec:previous work}
Existing floorplanning methods can be broadly divided into three categories: analytical methods, heuristic methods, and reinforcement learning (RL) methods.

Analytical methods, such as UFO~\cite{analytical-UFO} and PeF~\cite{analytical-pef}, typically operate in two phases: global distribution followed by legalization. In the global distribution phase, an initial rough floorplan with some overlaps is generated, which is then resolved during the legalization phase. While these methods are valued for their efficiency, they are limited by the requirement that the objective function be differentiable.

Heuristic methods depend on specific floorplan representations, such as the $B^{*}$-Tree~\cite{heuristic-SA} and sequence pair~\cite{NP-hard}. Additionally, the CB-Tree~\cite{heuiristic-CB} incorporates corner stitching constraints to address more complex requirements. These methods typically use simulated annealing to explore the solution space, often producing near-optimal solutions. However, they frequently require numerous iterations, reducing their computational efficiency.

Recently, reinforcement learning (RL) methods have gained attention as a promising approach to floorplanning, leveraging their ability to capture features from the chip canvas and improve performance through continuous environmental interaction. For example, GoodFloorplan~\cite{RL-goodFP} employs a graph convolutional network-based RL method, while~\cite{RL-CBL} uses a corner block list representation combined with hypergraph embedding. Nevertheless, these methods face challenges such as insufficient training data, limited model generalization, and the need for extensive iterative refinement.

\section{Preliminaries}\label{sec:preliminary}

\subsection{Floorplanning and Detailed Placement}
Floorplanning and detailed placement are the initial two stages of the physical design process. Given a netlist \((\mathcal{N}, \mathcal{M})\), where each net \(\mathcal{N}_i \in \mathcal{N}\) represents a connection relationship among modules and I/O ports, and each module \(M_i \in \mathcal{M}\) specifies its size \(a_i\) and the components \(m_i^j\) with areas \(a_i^j\) to be placed within \(M_i\), floorplanning seeks to optimally position and shape each circuit module \(M_i\) within a fixed chip canvas \((W, H)\)~\cite{TOFU}. Following floorplanning, the detailed placement stage aims to optimally position each component \(m_i^j\)—including larger macros and smaller standard cells—within each module \(M_i\). Both stages, constrained by conditions such as non-overlapping, strive to improve the chip’s overall Power, Performance, and Area (PPA) metrics. However, addressing the entire physical design flow and achieving final PPA metrics within a single algorithm is complex, so attention often shifts to a proxy metric, half-perimeter wire length (HPWL), defined as the sum of the half-perimeters of the bounding boxes for all nets in the netlist.

\subsection{Feedthrough}
Feedthrough is a routing technique that enables signals to pass through a module without being processed or modified by its internal logic. However, excessive use of feedthroughs can undermine module integrity and potentially cause blockages among internal components, complicating timing convergence and increasing design verification complexity. Ideally, modules connected within the same net should be adjacent in the final layout, allowing routing pins to be placed directly on their shared edges, thus minimizing disruptions to other modules from unnecessary feedthrough. Moreover, the length of the common edge determines the capacity for placing I/O pins between modules, ensuring sufficient spacing between pin pairs to meet practical industry constraints.

To effectively assess feedthrough, we introduce two metrics: the total number of feedthrough modules (FTmod) per net and the number of feedthrough pins (FTpin) on common edges. These are modeled as follows:
\begin{equation}
\label{eq:ftmod}
\text{FTmod}(\mathcal{N}_i) = \frac{1}{2}\sum_{M_j \in B_i} \mathbf{1}(M_j \text{ is a feedthrough module})
\end{equation}
\begin{equation}
\label{eq:ftpin}
\text{FTpin}(M_i, M_j) = \max\left(0, \left\lceil 
\frac{u \cdot Y_{ij} - CE_{ij}}{u} \right\rceil\right)
\end{equation}
where \(B_i\) represents the bounding box of net \(\mathcal{N}_i\), chosen to align with HPWL calculation methods; \(Y_{ij}\) denotes the total number of pins required between modules \(M_i\) and \(M_j\) as specified in the netlist; \(u\) is the minimum pin spacing; and \(CE_{ij}\) is the length of the common edge shared by modules \(M_i\) and \(M_j\). As illustrated in Figure~\ref{fig:motivation}, \(\text{FTmod}(\mathcal{N})\) equals 1 for \(\mathcal{N} = \{M_1, M_3\}\), indicating that \(M_2\) is the sole feedthrough module. Additionally, assume the common edge between \(M_1\) and \(M_2\) supports up to three I/O pins under the pin spacing constraint. If one pin is used for a feedthrough connection to \(M_3\) and three additional nets connect \(M_1\) and \(M_2\) (requiring three more pins, with only two available), then \(\text{FTpin}(M_1, M_2)\) equals 1.

\section{Proposed Method}\label{sec:propose method}
\subsection{Overview}
\begin{figure}[t]
    \centering
    \includegraphics[width=\linewidth]{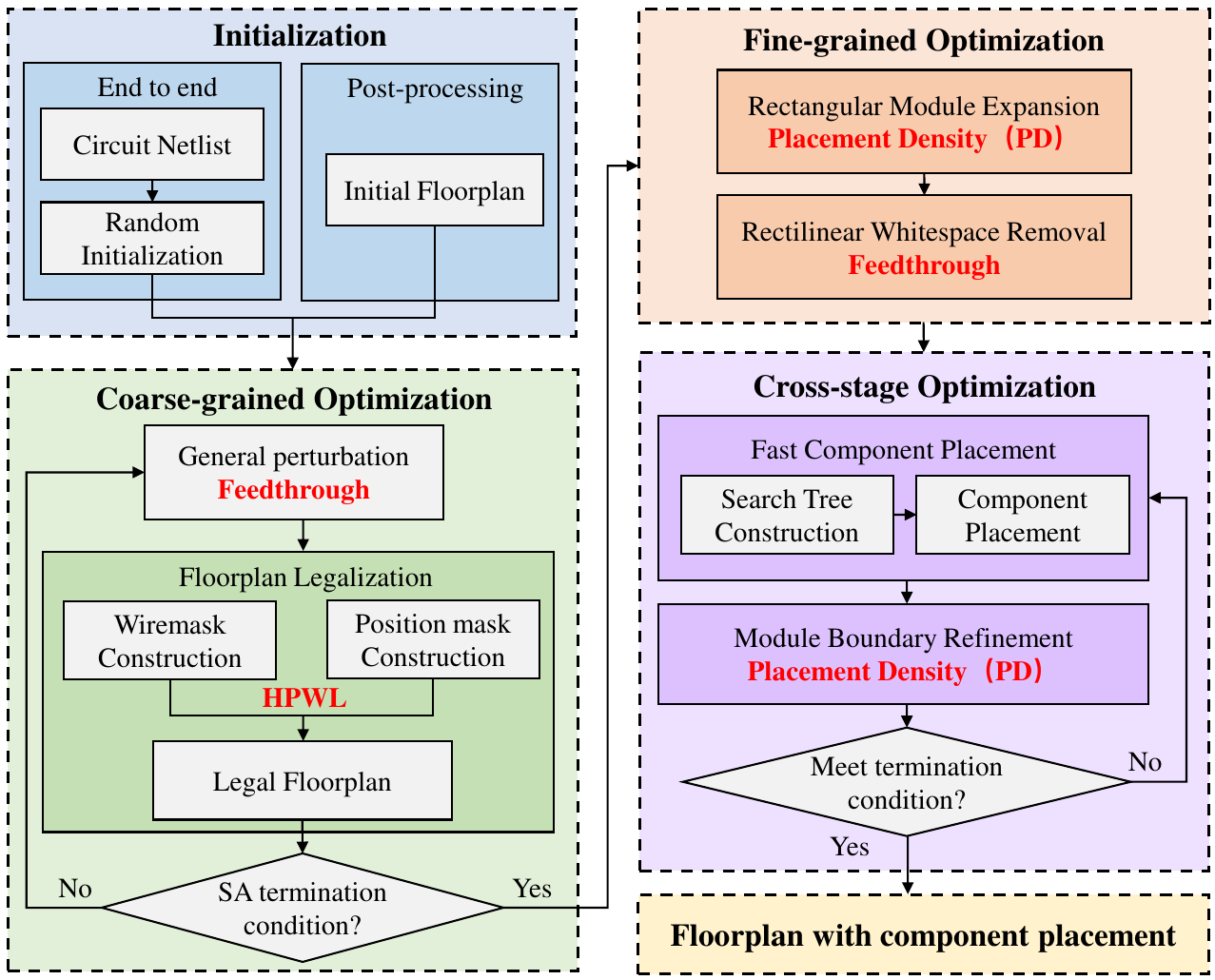}
    \caption{\textbf{Overall Flow of Flora.} The optimization metrics for each stage are highlighted in red.}
    \label{fig:overall flow}
\end{figure}

As shown in Figure~\ref{fig:overall flow}, Flora consists of three stages: (1) Coarse-grained optimization, (2) Fine-grained optimization, and (3) Cross-stage optimization. The details of our algorithm are discussed in Sections \ref{sec:floorplan legalization}, \ref{sec:rectilinear-based whitespace removal}, and \ref{sec:cross-stage-prediction}, respectively.

\subsection{Stage 1: Coarse-grained Optimization}\label{sec:floorplan legalization}
In the initial stage, our algorithm begins with an initialization. When Flora is used as a post-processing optimization for other algorithms, it starts from the final layout result of those algorithms. When Flora starts from scratch, it performs a random initialization. Subsequently, the layout undergoes coarse global optimization that effectively reduces both HPWL and feedthrough metrics under the constraint of non-overlapping modules.

We utilize the concepts of position mask and wiremask from Maskplace~\cite{maskplace} and Wiremask-EA~\cite{wiremask-bbo}. By dividing the chip canvas into discrete grids, the position mask identifies legal grids for placing the next module, while the wiremask indicates the HPWL increment associated with positioning the next module on each grid. Building on this insight, we introduce a feedthrough-aware simulated annealing (SA) algorithm. In each iteration, Flora randomly swaps two modules and legalizes the solution using the element-wise product of the wiremask and position mask, which identifies legal grids and their corresponding HPWL increments. Each illegal module is then moved sequentially to the legal grid with the smallest positive HPWL increment, prioritized in descending order of their initial areas \( a_i \), meaning larger modules are considered first. Furthermore, leveraging the wiremask’s inherent ability to optimize HPWL, we set the SA objective function as a weighted average of FTmod and FTpin, enabling joint optimization of feedthrough and HPWL.

\begin{figure}[t]
    \centering
    \includegraphics[width=\linewidth]{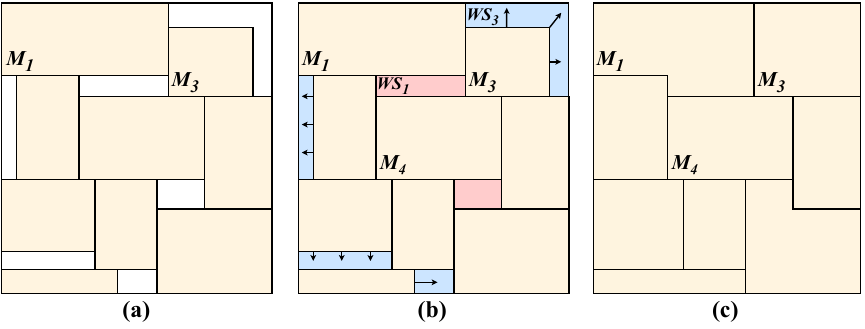}
    \caption{\textbf{Fine-grained Optimization.} (a) Illustrates the floorplan after Stage 1, (b) shows the rectangular module expansion, and (c) demonstrates the rectilinear whitespace removal, ultimately achieving a zero-whitespace layout.}
    \label{fig:whitespace}
\end{figure}

\subsection{Stage 2: Fine-grained Optimization}\label{sec:rectilinear-based whitespace removal}
In the second stage, Flora first sequentially expands the modules to improve component placement. It then allocates the remaining whitespace to further optimize feedthrough while maintaining relative module regularity. After these steps, Flora achieves a zero-whitespace layout.

\subsubsection{Rectangular Module Expansion}\label{sec:Rectangular Module Expansion}
In this step, we expand each module to enhance the potential for component placement. We first calculate the component area ratio as \( r_i = \frac{\sum_j a_i^j}{a_i} \). A smaller \( r_i \) indicates that placing components within module \( M_i \) is easier. Based on this insight, we dynamically allocate whitespace grids around module boundaries to each module in descending order of their \( r_i \) values, ensuring that each module retains a rectangular shape. This is because a rectangular shape facilitates better arrangement of larger internal macros. As illustrated in Figure~\ref{fig:whitespace}(b), each whitespace marked in blue is allocated to the corresponding module. Note that these four modules are the only ones that can maintain a rectangular shape after whitespace allocation, so the allocation result is the same regardless of the order of \( r_i \).

\begin{algorithm}[h]
\caption{Rectilinear Whitespace Removal}
\label{alg:whitespace_allocation}
\begin{algorithmic}[1]
\REQUIRE Rectangular layout \( \mathcal{C} \) with some whitespace
\ENSURE Zero-whitespace layout
\FOR{each empty grid \( g_{ij} \in \mathcal{C} \)}
    \STATE \( R_{ij} \gets \text{LargestBlankRectangle}(g_{ij}) \)
    \STATE \( \mathcal{L}_{ij} \gets \{ M_k \mid M_k \text{ is adjacent to } R_{ij} \} \)
    \STATE \( M^* \gets \arg\min_{M_k \in \mathcal{L}_{ij}} \Delta \text{FTpin}(R_{ij}, M_k) \)
    \STATE Allocate \( R_{ij} \) to \( M^* \)
\ENDFOR
\STATE \textbf{Function:} \(\Delta \text{FTpin}(R_{ij}, M_k)\)
\STATE \hspace{1em} \( \text{FTpin}_{\text{ori}} \gets \text{CalculateFTpin}(\mathcal{C}) \)
\STATE \hspace{1em} \( \mathcal{C}_{\text{orig}} \gets \mathcal{C} \)
\STATE \hspace{1em} \( \mathcal{C} \gets \text{AssignGrid}(\mathcal{C}, R_{ij}, M_k) \)
\STATE \hspace{1em} \( \text{FTpin}_{\text{new}} \gets \text{CalculateFTpin}(\mathcal{C}) \)
\STATE \hspace{1em} \( \mathcal{C} \gets \mathcal{C}_{\text{orig}} \)
\STATE \hspace{1em} \( \Delta \text{FTpin} \gets \text{FTpin}_{\text{new}} - \text{FTpin}_{\text{ori}} \)
\STATE \hspace{1em} \textbf{return} \( \Delta \text{FTpin} \)
\STATE \textbf{End Function}
\STATE \textbf{Return} Updated \( \mathcal{C} \)
\end{algorithmic}
\end{algorithm}

\subsubsection{Rectilinear Whitespace Removal}
After resizing each module, some whitespace remains due to the constraint of preserving rectangular shapes, as highlighted in red in Figure~\ref{fig:whitespace}(b). To address this and further improve chip utilization, we convert modules from rectangular to rectilinear shapes with the objective of minimizing feedthrough. The details of this method are outlined in Algorithm~\ref{alg:whitespace_allocation}. We iterate through each remaining blank grid, identifying the largest blank rectangle containing it. We then examine all modules adjacent to this blank rectangle and allocate it to the module that most effectively reduces the number of FTpins. This approach aims to increase the common edge length \( CE_{ij} \) between modules, allowing corresponding pins to be placed directly along their common edges. By allocating the largest blank rectangle, we ensure the relative regularity of the module layout. As shown in Figure~\ref{fig:whitespace}(c), if there is a net connection between $M_1$ and $M_4$, allocating \( WS_1 \) to module \( M_1 \) allows the pins of the two modules to be placed directly on their common edge boundary, eliminating the need for feedthrough.

\subsection{Stage 3: Cross-stage Optimization}\label{sec:cross-stage-prediction}
\begin{figure*}[t]
    \centering
    \includegraphics[width=\textwidth]{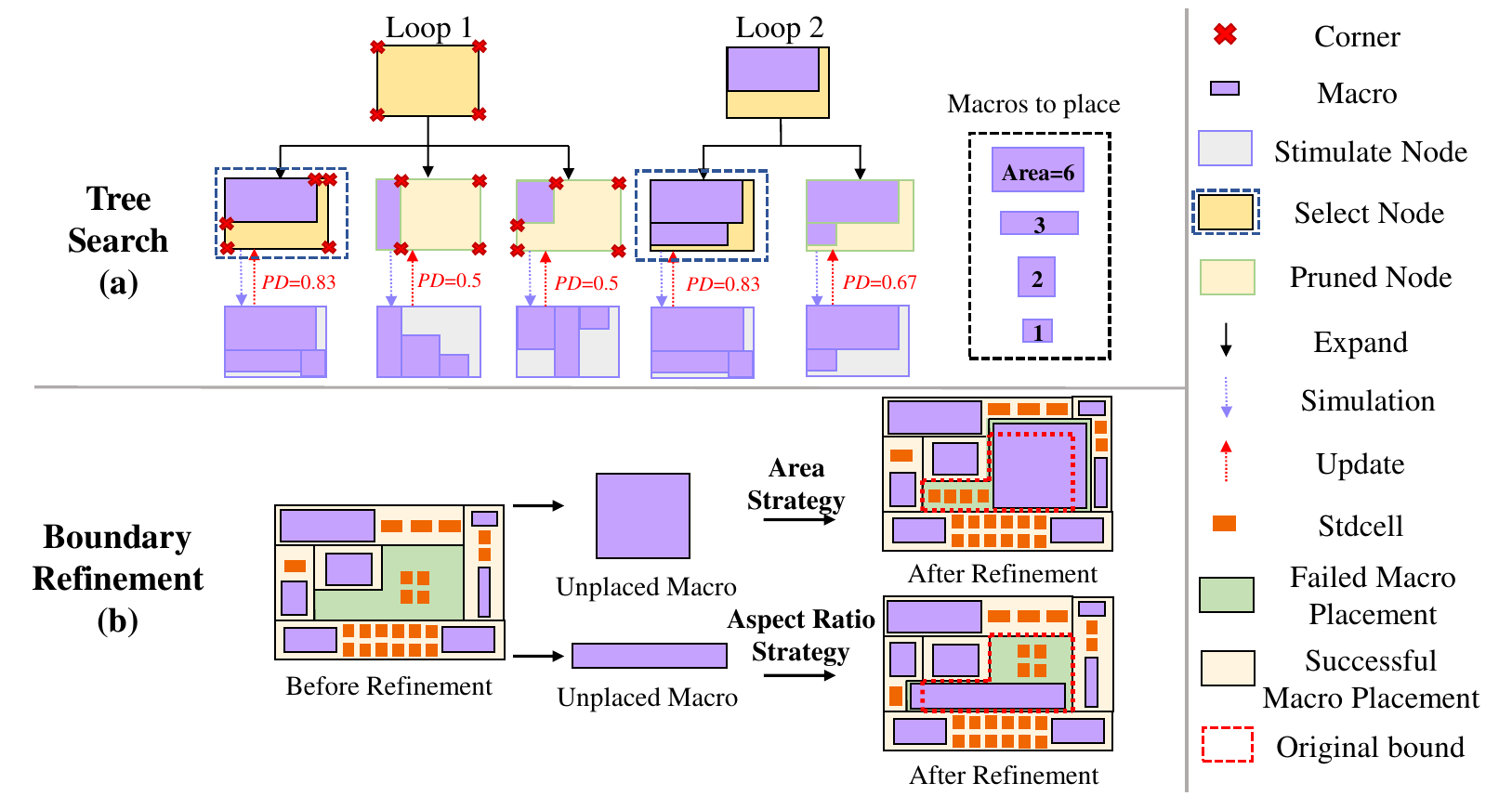}
    \caption{
    \textbf{Tree Search Method.} For each module, we construct a search tree where each node represents a partial placement. We ultimately place the components based on the node with the maximum PD value.
    \textbf{Boundary Refinement Method.} For unplaced macros within the module, we propose two strategies to enlarge the module for better placement.
    }
    \label{fig:macro}
\end{figure*}

In the third stage, we first introduce an algorithm for fast component placement, followed by a fine-tuning algorithm that dynamically adjusts module boundaries based on the placement results. The structure is demonstrated in Figure~\ref{fig:macro}.

\subsubsection{Fast Component Placement}
We begin by partitioning standard cells into multiple clusters~\cite{stdcell}, thereby reducing design complexity and allowing these clusters to be treated as special macros defined by their smallest bounding boxes. Using this partitioning and inspired by~\cite{2D_packing}, we construct a forest \( \mathcal{F} \), where each tree \( T_i \in \mathcal{F} \) corresponds to module \( M_i \), and each node \( N_{i}^c \in T_i \) represents a partial macro placement within \( M_i \).

The tree traversal begins with the root node \( N^{\text{root}}_i \), which corresponds to an empty module with no macros placed. In each iteration, we expand the current node \( N^*_i \) by generating all legal candidate nodes \( L_i = \{ N_i^1, N_i^2, \ldots, N_i^n \} \), each representing the placement of an unplaced macro \( m_i^k \) at an empty corner of the current canvas corresponding to \( N^*_i \). Thus, each candidate node extends the macro placement of \( N^*_i \) by legally adding one more macro at a corner of the current module. We constrain each macro to be placed at a corner. While this assumption may not fully reflect real-world placement strategies, it allows for rapid evaluation of the worst-case scenario—whether the module can accommodate all components \( m_i^j \). By placing each macro as close to the corners as possible, we facilitate the placement of more and larger macros within the module.

Subsequently, we evaluate each candidate node using the placement density (PD) to measure the area of placed macros, calculated as:
\begin{equation}
\label{eq:PD}
    \text{PD}(M_i) = \frac{\sum_{j} a_i^j \cdot \mathbf{1}(\text{success}(m_i^j))}{\sum_{j} a_i^j}
\end{equation}
Our goal is for each tree \( T_i \) to have a final node with a PD value of 1, indicating that all components within the module can be properly placed.

\begin{algorithm}[H]
\caption{Fast Component Placement}
\label{alg:Fast Placement}
\begin{algorithmic}[1]
\REQUIRE Initial layout \( \mathcal{C} \)
\ENSURE Layout with component placement
\STATE \(\text{PartitionClusters}(\mathcal{M})\)
\STATE \( \mathcal{F} \gets \text{ConstructEmptyForest}(\mathcal{M}) \)
\FOR{each tree \( T_i \in \mathcal{F} \)}
    \STATE Initialize \( L_i \) with the root node \( N_i^{\text{root}} \)
    \WHILE{\( L_i \neq \emptyset \)}
        \FOR{each node \( N_i^c \in L_i \)}
            \STATE \( N_i^c.\text{value} \gets \text{Simulation}(N_i^c) \)
            \IF{\( N_i^c.\text{value} == 1 \)}
                \STATE \textbf{return} \( \mathcal{C} \)
            \ENDIF
        \ENDFOR
        \STATE \( N^*_i \gets \arg\max_{N_i^c \in L_i} N_i^c.\text{value} \)
        \STATE \( L_i \gets \text{Expand}(N^*_i) \)
    \ENDWHILE
    \STATE Update the component placement of \( M_i \) with \( N^*_i \)
\ENDFOR
\STATE \textbf{Return} \( \mathcal{C} \)
\STATE \textbf{Function:} \(\text{Simulation}(N_i^c)\)
\STATE \hspace{1em} \( \mathcal{S}_i \gets \{ \text{remaining macros} \} \)
\STATE \hspace{1em} Sort \( \mathcal{S}_i \) in descending order of macro area
\STATE \hspace{1em} \textbf{while} \( \mathcal{S}_i \) is not empty \textbf{do}
\STATE \hspace{2em} \( m \gets \mathcal{S}_i.\text{pop}() \)
\STATE \hspace{2em} Find all corners \( L = \{C_1, C_2, \ldots, C_n\} \) to place \( m \)
\STATE \hspace{2em} \textbf{if} \( L \) is empty \textbf{then}
\STATE \hspace{3em} continue
\STATE \hspace{2em} \textbf{else}
\STATE \hspace{3em} \( C^* \gets \arg\min_{C_i} \Delta \text{Cor}_{\text{num}} \)
\STATE \hspace{3em} \(\text{Place}(m, C^*)\)
\STATE \hspace{1em} \textbf{end while}
\STATE \hspace{1em} \textbf{return} \( \sum_{j} a_i^j \cdot \mathbf{1}(\text{success}(m_i^j)) / {\sum_{j} a_i^j} \)
\STATE \textbf{End Function}
\end{algorithmic}
\end{algorithm}

To this end, for each candidate node \( N_i^c \), we first perform a fast simulation to estimate its potential PD value. The simulation steps are as follows: we sort the remaining unplaced macros in descending order of their areas to obtain \( \mathcal{S}_i \). In each step, we select a macro \( m \) from \( \mathcal{S}_i \) and attempt to place it at a corner of the remaining available space in the module. If no corner can accommodate \( m \), we proceed to the next macro in \( \mathcal{S}_i \). If multiple corners can legally place \( m \), we choose the corner \( C^* \) that minimizes the number of corners \( \text{Cor}_{\text{num}} \) in the remaining available space after placement. This is because, regardless of where \( m \) is placed, the remaining area is the same, but a layout with fewer corners indicates a more regular remaining space, which is more conducive to placing subsequent macros. This process iterates until no remaining macros can be properly placed. We then update the PD value of \( N_i^c \) based on the final placement density.

After simulating each candidate node, we select the node \( N^*_i \) with the highest PD value and proceed to the next iteration. The details of the algorithm are provided in Algorithm~\ref{alg:Fast Placement}.

\subsubsection{Module Boundary Refinement}
Ideally, the PD value of the final node \( N^*_i \) for each module would be 1, indicating that the module’s area and shape can effectively accommodate all internal components. However, this is often not achievable in a single step. As illustrated in Figure~\ref{fig:macro}(b), some modules cannot accommodate all internal macros due to limited area or irregular shapes. We categorize these unplaced macros into two groups: those with large areas (aspect ratio \(\leq 2\)) and those with high aspect ratios (aspect ratio \(> 2\)). To address this, we developed a refinement algorithm with two strategies.

For each module \( M_i \) with a PD value less than 1, let us assume the unplaced macro is \( m_i \). If there are multiple unplaced macros, we iteratively apply the following method. First, we compare the aspect ratio of \( m_i \) with the largest-area rectangle and the rectangle with the highest aspect ratio in the module’s available space, then select the rectangle whose aspect ratio best matches that of the macro, as shown in Figure~\ref{fig:macro}(b).

Next, we dynamically expand this rectangle by allocating grids from adjacent modules until the macro can be properly placed or until further expansion would interfere with the placement of surrounding modules. Each expansion proceeds grid by grid in the direction of the adjacent module with the smallest component area ratio \( r_i \) (as defined in Section \ref{sec:Rectangular Module Expansion}), aiming to allocate grids from modules with a PD value of 1 and sufficient remaining space to adjacent modules with a PD value less than 1, thereby increasing their PD value.

\newcolumntype{C}{>{\centering\arraybackslash}X}

\begin{table*}[t]
\centering
\caption{\textbf{Main Results}. The main results for the HPWL, FTpin (feedthrough pin), FTmod (feedthrough module), WS (whitespace), PD (placement density), and RT(runtime seconds) metrics on the GSRC and MCNC benchmarks. $\uparrow$ indicates that a larger value is better. The \textbf{\textcolor{deepblue}{best}} results are highlighted in bold red, and the \textcolor{color2}{\underline{second-best}} are underlined in blue.}
\label{tab:main}

\begin{tabularx}{\textwidth}{CCCCCCCCCC} 
\toprule
\textbf{Method} &
  \textbf{Metric} &
  \textbf{n10} &
  \textbf{n30} &
  \textbf{n50} &
  \textbf{n100} &
  \textbf{n200} &
  \textbf{n300} &
  \textbf{ami33} &
  \textbf{ami49} \\ 
\midrule
Corblivar &
  \begin{tabular}[c]{@{}c@{}} 
    HPWL  \\ 
    FTpin  \\ 
    FTmod  \\ 
    WS(\%)  \\ 
    PD(\%) $\uparrow$ \\
    RT(s) 
  \end{tabular} &
  \begin{tabular}[c]{@{}c@{}}46524\\ 53\\ 256\\ 13.59\\ \textcolor{color2}{\underline{88.94}}\\ \textbf{\textcolor{deepblue}{0.05}}\end{tabular} & 
  \begin{tabular}[c]{@{}c@{}}118209\\ 162\\ 1221.5\\ 11.81\\ 82.14\\ \textbf{\textcolor{deepblue}{0.37}}\end{tabular} & 
  \begin{tabular}[c]{@{}c@{}}149215\\ \textcolor{color2}{\underline{365}}\\ 1990.5\\ 13.93\\ \textcolor{color2}{\underline{86.76}}\\ \textbf{\textcolor{deepblue}{1.03}}\end{tabular} & 
  \begin{tabular}[c]{@{}c@{}}258928\\ 707\\ 5756\\ 15.72\\ 87.23\\ \textcolor{color2}{\underline{5.13}}\end{tabular} & 
  \begin{tabular}[c]{@{}c@{}}519523\\ 1599\\ 16912.5\\ 14.85\\ 79.79\\ \textcolor{color2}{\underline{28.24}}\end{tabular} & 
  \begin{tabular}[c]{@{}c@{}}695832\\ 2121\\ 22552\\ 16.50\\ 82.31\\ \textcolor{color2}{\underline{68.01}}\end{tabular} & 
  \begin{tabular}[c]{@{}c@{}}84369\\ 534\\ 327\\ 12.25\\ 84.66\\ \textbf{\textcolor{deepblue}{0.34}}\end{tabular} & 
  \begin{tabular}[c]{@{}c@{}}1535641\\ 719\\ 1987.5\\ 19.10\\ 79.41\\ \textbf{\textcolor{deepblue}{0.82}}\end{tabular} \\ 
\midrule
TOFU &
  \begin{tabular}[c]{@{}c@{}}
    HPWL  \\ 
    FTpin  \\ 
    FTmod  \\ 
    WS(\%)  \\ 
    PD(\%) $\uparrow$ \\
    RT(s) 
  \end{tabular} &
  \begin{tabular}[c]{@{}c@{}} \textcolor{color2}{\underline{35213}}\\ \textcolor{color2}{\underline{52}}\\ \textcolor{color2}{\underline{124}}\\ \textcolor{color2}{\underline{1.52}}\\ 85.29\\ \textcolor{color2}{\underline{1.35}}\end{tabular} & 
  \begin{tabular}[c]{@{}c@{}}\textcolor{color2}{\underline{109189}}\\ \textcolor{color2}{\underline{159}}\\ \textcolor{color2}{\underline{898}}\\ \textcolor{color2}{\underline{0.61}}\\ \textcolor{color2}{\underline{87.48}}\\ \textcolor{color2}{\underline{0.97}}\end{tabular} & 
  \begin{tabular}[c]{@{}c@{}}\textcolor{color2}{\underline{147650}}\\ 367\\ \textcolor{color2}{\underline{1750}}\\ \textcolor{color2}{\underline{0.56}}\\ 83.65\\ \textcolor{color2}{\underline{1.55}}\end{tabular} & 
  \begin{tabular}[c]{@{}c@{}}250942\\ 702\\ \textcolor{color2}{\underline{5160}}\\ \textcolor{color2}{\underline{1.22}}\\ \textcolor{color2}{\underline{90.80}}\\ \textbf{\textcolor{deepblue}{2.58}}\end{tabular} & 
  \begin{tabular}[c]{@{}c@{}}484507\\ 1590\\ 15660\\ \textcolor{color2}{\underline{1.57}}\\ \textcolor{color2}{\underline{83.98}}\\ \textbf{\textcolor{deepblue}{3.81}}\end{tabular} & 
  \begin{tabular}[c]{@{}c@{}}712547\\ 2124\\ \textcolor{color2}{\underline{20780}}\\ \textcolor{color2}{\underline{1.54}}\\ 83.66\\ \textbf{\textcolor{deepblue}{8.21}}\end{tabular} & 
  \begin{tabular}[c]{@{}c@{}}\textcolor{color2}{\underline{71069}}\\ \textcolor{color2}{\underline{524}}\\ \textcolor{color2}{\underline{255.5}}\\ \textcolor{color2}{\underline{0.63}}\\ \textcolor{color2}{\underline{88.74}}\\ \textcolor{color2}{\underline{1.15}}\end{tabular} & 
  \begin{tabular}[c]{@{}c@{}}1144620\\ \textcolor{color2}{\underline{708}}\\ 1654\\ \textcolor{color2}{\underline{0.56}}\\ \textcolor{color2}{\underline{79.43}}\\ \textcolor{color2}{\underline{1.58}}\end{tabular} \\ 
\midrule
Wiremask-EA &
  \begin{tabular}[c]{@{}c@{}}
    HPWL  \\ 
    FTpin  \\ 
    FTmod  \\ 
    WS(\%)  \\ 
    PD(\%) $\uparrow$ \\
    RT(s) 
  \end{tabular} &
  \begin{tabular}[c]{@{}c@{}}36669\\ 54\\ 260\\ 23.91\\ 84.95\\ 27.71\end{tabular} & 
  \begin{tabular}[c]{@{}c@{}}116688\\ 163\\ 1419.5\\ 15.97\\ 82.44\\ 57.84\end{tabular} & 
  \begin{tabular}[c]{@{}c@{}}149710\\ 368\\ 1938\\ 8.29\\ 80.58\\ 83.70\end{tabular} & 
  \begin{tabular}[c]{@{}c@{}}\textcolor{color2}{\underline{236714}}\\ \textcolor{color2}{\underline{698}}\\ 5317\\ 9.57\\ 87.65\\ 124.58\end{tabular} & 
  \begin{tabular}[c]{@{}c@{}}\textcolor{color2}{\underline{471472}}\\ \textcolor{color2}{\underline{1549}}\\ \textcolor{color2}{\underline{15021}}\\ 9.60\\ 83.92\\ 210.28\end{tabular} & 
  \begin{tabular}[c]{@{}c@{}}\textcolor{color2}{\underline{675116}}\\ \textcolor{color2}{\underline{2088}}\\ 21377.5\\ 14.67\\ \textcolor{color2}{\underline{87.01}}\\ 283.12\end{tabular} & 
  \begin{tabular}[c]{@{}c@{}}71631\\ 545\\ 333\\ 25.58\\ 64.86\\ 57.34\end{tabular} & 
  \begin{tabular}[c]{@{}c@{}}\textcolor{color2}{\underline{965377}}\\ 713\\ \textcolor{color2}{\underline{1319.5}}\\ 20.98\\ 76.98\\ 78.26\end{tabular} \\ 
\midrule
Flora &
  \begin{tabular}[c]{@{}c@{}}
    HPWL \\ 
    FTpin \\ 
    FTmod \\ 
    WS(\%) \\ 
    PD(\%) $\uparrow$ \\
    RT(s) 
  \end{tabular} &
  \begin{tabular}[c]{@{}c@{}}
    \textbf{\textcolor{deepblue}{34742}}\\ 
    \textbf{\textcolor{deepblue}{43}}\\ 
    \textbf{\textcolor{deepblue}{89.5}}\\ 
    \textbf{\textcolor{deepblue}{0}}\\ 
    \textbf{\textcolor{deepblue}{100}}\\ 
    22.89
  \end{tabular} & 
  \begin{tabular}[c]{@{}c@{}}
    \textbf{\textcolor{deepblue}{107340}}\\ 
    \textbf{\textcolor{deepblue}{146}}\\ 
    \textbf{\textcolor{deepblue}{789}}\\ 
    \textbf{\textcolor{deepblue}{0}}\\ 
    \textbf{\textcolor{deepblue}{100}}\\ 
    59.84
  \end{tabular} & 
  \begin{tabular}[c]{@{}c@{}}
    \textbf{\textcolor{deepblue}{144299}}\\ 
    \textbf{\textcolor{deepblue}{354}}\\ 
    \textbf{\textcolor{deepblue}{1469.5}}\\ 
    \textbf{\textcolor{deepblue}{0}}\\ 
    \textbf{\textcolor{deepblue}{99.72}}\\ 
    94.64
  \end{tabular} & 
  \begin{tabular}[c]{@{}c@{}}
    \textbf{\textcolor{deepblue}{226705}}\\ 
    \textbf{\textcolor{deepblue}{681}}\\ 
    \textbf{\textcolor{deepblue}{3826}}\\ 
    \textbf{\textcolor{deepblue}{0}}\\ 
    \textbf{\textcolor{deepblue}{100}}\\ 
    154.33
  \end{tabular} & 
  \begin{tabular}[c]{@{}c@{}}
    \textbf{\textcolor{deepblue}{465399}}\\ 
    \textbf{\textcolor{deepblue}{1534}}\\ 
    \textbf{\textcolor{deepblue}{10918.5}}\\ 
    \textbf{\textcolor{deepblue}{0}}\\ 
    \textbf{\textcolor{deepblue}{100}}\\ 
    310.41
  \end{tabular} & 
  \begin{tabular}[c]{@{}c@{}}
    \textbf{\textcolor{deepblue}{597486}}\\ 
    \textbf{\textcolor{deepblue}{2027}}\\ 
    \textbf{\textcolor{deepblue}{15626}}\\ 
    \textbf{\textcolor{deepblue}{0}}\\ 
    \textbf{\textcolor{deepblue}{97.83}}\\ 
    441.69
  \end{tabular} & 
  \begin{tabular}[c]{@{}c@{}}
    \textbf{\textcolor{deepblue}{63976}}\\ 
    \textbf{\textcolor{deepblue}{516}}\\ 
    \textbf{\textcolor{deepblue}{145}}\\ 
    \textbf{\textcolor{deepblue}{0}}\\ 
    \textbf{\textcolor{deepblue}{92.95}}\\ 
    54.48
  \end{tabular} & 
  \begin{tabular}[c]{@{}c@{}}
    \textbf{\textcolor{deepblue}{819560}}\\ 
    \textbf{\textcolor{deepblue}{673}}\\ 
    \textbf{\textcolor{deepblue}{580.5}}\\ 
    \textbf{\textcolor{deepblue}{0}}\\ 
    \textbf{\textcolor{deepblue}{100}}\\ 
    111.04
  \end{tabular} \\ 
\bottomrule
\end{tabularx}
\end{table*}

\begin{figure*}[!t]
    \centering
    \begin{subfigure}{\textwidth}
        \centering
        \begin{subfigure}[t]{0.24\textwidth}
            \centering
            \includegraphics[width=\linewidth, page=1]{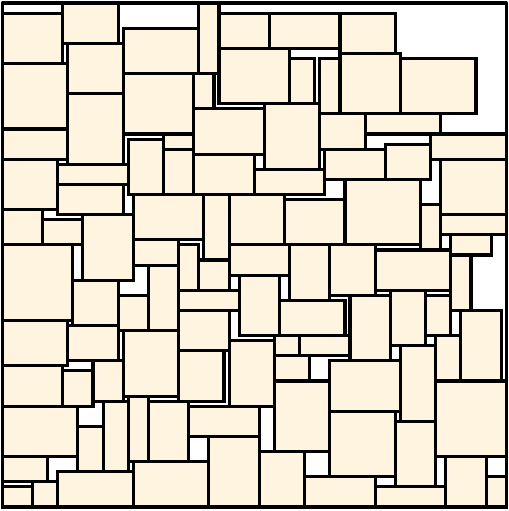}
            \caption{Corblivar}
            \label{fig:total_show_page1}
        \end{subfigure}
        \hfill
        \begin{subfigure}[t]{0.24\textwidth}
            \centering
            \includegraphics[width=\linewidth, page=2]{total_show.pdf}
            \caption{Stage1}
            \label{fig:total_show_page2}
        \end{subfigure}
        \hfill
        \begin{subfigure}[t]{0.24\textwidth}
            \centering
            \includegraphics[width=\linewidth, page=3]{total_show.pdf}
            \caption{Stage1+2}
            \label{fig:total_show_page3}
        \end{subfigure}
        \hfill
        \begin{subfigure}[t]{0.24\textwidth}
            \centering
            \includegraphics[width=\linewidth, page=4]{total_show.pdf}
            \caption{Stage1+2+3}
            \label{fig:total_show_page4}
        \end{subfigure}
    \end{subfigure}
    \caption{\textbf{Floorplan Example.} This figure displays the layout results of the n100 designs using Corblivar and Flora. Additionally, it showcases the results for each of Flora's three stages. Macros and standard cell clusters are colored in purple and red, respectively.}
    \label{fig:total_show}
\end{figure*}

\begin{table*}[!t]
\centering
\caption{\textbf{Ablation Study Results.} Performance of each stage of Flora as a post-processing algorithm for Corblivar. $\Delta \text{RT(s)}$ represents the runtime of the current stage.}
\label{tab:ablation}
\begin{tabularx}{\textwidth}{CCCCCCCCCC} 
\toprule
\textbf{Method} & \textbf{Metric} & \textbf{n10} & \textbf{n30} & \textbf{n50} & \textbf{n100} & \textbf{n200} & \textbf{n300} & \textbf{ami33} & \textbf{ami49} \\ 
\midrule
Corblivar & 
  \begin{tabular}[c]{@{}c@{}}HPWL  \\ FTpin  \\ FTmod  \\ WS(\%)  \\ PD(\%) $\uparrow$ \\ $\Delta \text{RT(s)}$ \end{tabular} &
\begin{tabular}[c]{@{}c@{}}46524\\ 53\\ 256\\ \textcolor{color2}{\underline{13.59}}\\ 88.94\\ \textbf{\textcolor{deepblue}{0.05}}\end{tabular} & 
\begin{tabular}[c]{@{}c@{}}118209\\ 162\\ 1221.5\\ \textcolor{color2}{\underline{11.81}}\\ 82.14\\ \textbf{\textcolor{deepblue}{0.37}}\end{tabular} & 
\begin{tabular}[c]{@{}c@{}}149215\\ 365\\ 1990.5\\ \textcolor{color2}{\underline{13.93}}\\ 86.76\\ \textcolor{color2}{\underline{1.03}}\end{tabular} & 
\begin{tabular}[c]{@{}c@{}}258928\\ 707\\ 5756\\ 15.72\\ 87.23\\ \textcolor{color2}{\underline{5.13}}\end{tabular} & 
\begin{tabular}[c]{@{}c@{}}519523\\ 1599\\ 16912.5\\ \textcolor{color2}{\underline{14.85}}\\ 79.79\\ \textcolor{color2}{\underline{28.24}}\end{tabular} & 
\begin{tabular}[c]{@{}c@{}}695832\\ 2121\\ 22552\\ 16.50\\ 82.31\\ \textcolor{color2}{\underline{68.01}}\end{tabular} & 
\begin{tabular}[c]{@{}c@{}}84369\\ 534\\ 327\\ \textcolor{color2}{\underline{12.25}}\\ \textcolor{color2}{\underline{84.66}}\\ \textbf{\textcolor{deepblue}{0.34}}\end{tabular} & 
\begin{tabular}[c]{@{}c@{}}1535641\\ 719\\ 1987.5\\ \textcolor{color2}{\underline{19.10}}\\ 79.41\\ \textcolor{color2}{\underline{0.82}}\end{tabular} \\ 
\midrule
Stage1 & 
  \begin{tabular}[c]{@{}c@{}}HPWL  \\ FTpin  \\ FTmod  \\ WS(\%)  \\ PD(\%) $\uparrow$ \\ $\Delta \text{RT(s)}$ \end{tabular} &
\begin{tabular}[c]{@{}c@{}} 
  \textbf{\textcolor{deepblue}{39252}}\\ 
  49\\ 
  242.5\\ 
  33.04\\ 
  72.27\\ 
  22.09
\end{tabular} & 
\begin{tabular}[c]{@{}c@{}} 
  \textbf{\textcolor{deepblue}{112592}}\\ 
  159\\ 
  1195.5\\ 
  19.49\\ 
  77.80\\ 
  30.10
\end{tabular} & 
\begin{tabular}[c]{@{}c@{}} 
  \textbf{\textcolor{deepblue}{143533}}\\ 
  \textcolor{color2}{\underline{356}}\\ 
  1924\\ 
  17.65\\ 
  84.75\\ 
  34.05
\end{tabular} & 
\begin{tabular}[c]{@{}c@{}} 
  \textbf{\textcolor{deepblue}{237430}}\\ 
  690\\ 
  5001\\ 
  \textcolor{color2}{\underline{14.50}}\\ 
  88.86\\ 
  49.88
\end{tabular} & 
\begin{tabular}[c]{@{}c@{}} 
  \textbf{\textcolor{deepblue}{439983}}\\ 
  1556\\ 
  13806\\ 
  15.13\\ 
  84.98\\ 
  84.33
\end{tabular} & 
\begin{tabular}[c]{@{}c@{}} 
  \textbf{\textcolor{deepblue}{632881}}\\ 
  2077\\ 
  19129.5\\ 
  \textcolor{color2}{\underline{12.87}}\\ 
  88.27\\ 
  115.72
\end{tabular} & 
\begin{tabular}[c]{@{}c@{}} 
  \textbf{\textcolor{deepblue}{63990}}\\ 
  530\\ 
  271\\ 
  33.12\\ 
  61.28\\ 
  30.11
\end{tabular} & 
\begin{tabular}[c]{@{}c@{}} 
  \textbf{\textcolor{deepblue}{952018}}\\ 
  705\\ 
  1039.5\\ 
  21.98\\ 
  80.19\\ 
  30.99
\end{tabular} \\ 
\midrule
Stage1+2 & 
  \begin{tabular}[c]{@{}c@{}}HPWL  \\ FTpin  \\ FTmod  \\ WS(\%)  \\ PD(\%) $\uparrow$ \\ $\Delta \text{RT(s)}$ \end{tabular} &
\begin{tabular}[c]{@{}c@{}}\textcolor{color2}{\underline{40870}}\\ \textcolor{color2}{\underline{42}}\\ \textbf{\textcolor{deepblue}{118}}\\ \textbf{\textcolor{deepblue}{0}}\\ \textcolor{color2}{\underline{91.99}}\\ \textcolor{color2}{\underline{0.36}}\end{tabular} & 
\begin{tabular}[c]{@{}c@{}}\textcolor{color2}{\underline{114377}}\\ \textcolor{color2}{\underline{153}}\\ \textbf{\textcolor{deepblue}{855.5}}\\ \textbf{\textcolor{deepblue}{0}}\\ \textcolor{color2}{\underline{87.07}}\\ \textcolor{color2}{\underline{0.46}}\end{tabular} & 
\begin{tabular}[c]{@{}c@{}}\textcolor{color2}{\underline{145590}}\\ \textbf{\textcolor{deepblue}{344}}\\ \textbf{\textcolor{deepblue}{1507}}\\ \textbf{\textcolor{deepblue}{0}}\\ \textcolor{color2}{\underline{90.15}}\\ \textbf{\textcolor{deepblue}{0.49}}\end{tabular} & 
\begin{tabular}[c]{@{}c@{}}\textcolor{color2}{\underline{239531}}\\ \textcolor{color2}{\underline{680}}\\ \textbf{\textcolor{deepblue}{3958.5}}\\ \textbf{\textcolor{deepblue}{0}}\\ \textcolor{color2}{\underline{93.35}}\\ \textbf{\textcolor{deepblue}{0.64}}\end{tabular} & 
\begin{tabular}[c]{@{}c@{}}\textcolor{color2}{\underline{442724}}\\ \textcolor{color2}{\underline{1532}}\\ \textbf{\textcolor{deepblue}{11677.5}}\\ \textbf{\textcolor{deepblue}{0}}\\ \textcolor{color2}{\underline{90.10}}\\ \textbf{\textcolor{deepblue}{0.89}}\end{tabular} & 
\begin{tabular}[c]{@{}c@{}}\textcolor{color2}{\underline{635427}}\\ \textcolor{color2}{\underline{2057}}\\ \textbf{\textcolor{deepblue}{16893}}\\ \textbf{\textcolor{deepblue}{0}}\\ \textcolor{color2}{\underline{92.75}}\\ \textbf{\textcolor{deepblue}{1.10}}\end{tabular} & 
\begin{tabular}[c]{@{}c@{}}\textcolor{color2}{\underline{69069}}\\ \textbf{\textcolor{deepblue}{524}}\\ \textbf{\textcolor{deepblue}{156.5}}\\ \textbf{\textcolor{deepblue}{0}}\\ 77.48\\ \textcolor{color2}{\underline{0.49}}\end{tabular} & 
\begin{tabular}[c]{@{}c@{}}\textcolor{color2}{\underline{996544}}\\ \textbf{\textcolor{deepblue}{682}}\\ \textbf{\textcolor{deepblue}{738.5}}\\ \textbf{\textcolor{deepblue}{0}}\\ \textcolor{color2}{\underline{85.91}}\\ \textbf{\textcolor{deepblue}{0.41}}\end{tabular} \\ 
\midrule
Stage1+2+3 & 
  \begin{tabular}[c]{@{}c@{}}HPWL  \\ FTpin  \\ FTmod  \\ WS(\%)  \\ PD(\%) $\uparrow$ \\ $\Delta \text{RT(s)}$ \end{tabular} & 
\begin{tabular}[c]{@{}c@{}}41236\\ \textbf{\textcolor{deepblue}{41}}\\ \textcolor{color2}{\underline{123.5}}\\ \textbf{\textcolor{deepblue}{0}}\\ \textbf{\textcolor{deepblue}{100}}\\ 10.69\end{tabular} & 
\begin{tabular}[c]{@{}c@{}}116343\\ \textbf{\textcolor{deepblue}{149}}\\ \textcolor{color2}{\underline{860.5}}\\ \textbf{\textcolor{deepblue}{0}}\\ \textbf{\textcolor{deepblue}{94.29}}\\ 33.34\end{tabular} & 
\begin{tabular}[c]{@{}c@{}}147215\\ \textbf{\textcolor{deepblue}{344}}\\ \textcolor{color2}{\underline{1528}}\\ \textbf{\textcolor{deepblue}{0}}\\ \textbf{\textcolor{deepblue}{99.92}}\\ 86.01\end{tabular} & 
\begin{tabular}[c]{@{}c@{}}241337\\ \textbf{\textcolor{deepblue}{671}}\\ \textcolor{color2}{\underline{4027.5}}\\ \textbf{\textcolor{deepblue}{0}}\\ \textbf{\textcolor{deepblue}{97.56}}\\ 135.20\end{tabular} & 
\begin{tabular}[c]{@{}c@{}}445666\\ \textbf{\textcolor{deepblue}{1526}}\\ \textcolor{color2}{\underline{12028}}\\ \textbf{\textcolor{deepblue}{0}}\\ \textbf{\textcolor{deepblue}{94.60}}\\ 267.22\end{tabular} & 
\begin{tabular}[c]{@{}c@{}}637986\\ \textbf{\textcolor{deepblue}{2056}}\\ \textcolor{color2}{\underline{17405.5}}\\ \textbf{\textcolor{deepblue}{0}}\\ \textbf{\textcolor{deepblue}{95.84}}\\ 383.42\end{tabular} & 
\begin{tabular}[c]{@{}c@{}}71462\\ \textcolor{color2}{\underline{525}}\\ \textcolor{color2}{\underline{165.5}}\\ \textbf{\textcolor{deepblue}{0}}\\ \textbf{\textcolor{deepblue}{92.92}}\\ 33.95\end{tabular} & 
\begin{tabular}[c]{@{}c@{}}1036689\\ \textcolor{color2}{\underline{683}}\\ \textcolor{color2}{\underline{772.5}}\\ \textbf{\textcolor{deepblue}{0}}\\ \textbf{\textcolor{deepblue}{97.91}}\\ 99.40\end{tabular} \\
\bottomrule
\end{tabularx}
\end{table*}

\section{Experimental Results}\label{sec:experiment}
\subsection{Benchmarks and Settings}
In this section, we evaluate the performance of Flora. All experiments are conducted on a single machine equipped with Intel Xeon E5-2667 v4 CPUs operating at 3.20 GHz.

For benchmarking, we employ the widely recognized MCNC~\cite{mcnc} and GSRC~\cite{gsrc} datasets, which feature module counts ranging from 10 to 300. As these benchmarks lack details on the internal macros and standard cells within modules, we manually configure the total components of each module to occupy 80\% of its area, aligning more closely with real-world chip design practices. We further assume that standard cells form clusters, treating them as a specialized type of macro. The number of macros per module is uniformly distributed between 2 and 6.

We compare Flora against three baseline algorithms: Corblivar~\cite{corb}, TOFU~\cite{TOFU}, and Wiremask-EA~\cite{wiremask-bbo}. Corblivar, a highly efficient simulated annealing (SA) method, ranks among the top open-source floorplanning algorithms. TOFU, a rectilinear framework, reduces whitespace by approximately 70\%. Wiremask-EA excels in minimizing HPWL. For evaluation, we adopt six metrics: HPWL, FTmod (Equation \ref{eq:ftmod}), FTpin (Equation \ref{eq:ftpin}), remaining whitespace area across the canvas (WS), placement density (PD, Equation \ref{eq:PD}), and runtime (RT).

\textbf{Model Settings:} The grid size for both Flora and Wiremask-EA is set to \(224 \times 224\). In the first stage of our algorithm, the SA strategy uses the following hyperparameters: an initial temperature \( T_{\text{init}} = 2000 \), a final temperature \( T_{\text{end}} = 1 \times 10^{-3} \), and a cooling rate of 0.99. Since TOFU is not open-source, we assess our metrics using solutions provided by its authors. Each method is evaluated over five independent runs, with mean values reported.

\subsection{Main Results}
Table \ref{tab:main} summarizes the performance of various floorplanning methods. Flora is implemented from scratch in these experiments. The results demonstrate that our algorithm surpasses state-of-the-art (SOTA) methods in five of the six metrics across eight datasets, with runtime (RT) as the sole exception. Specifically, Flora achieves average reductions of 5.93\% in HPWL, 5.16\% in FTpin, and 29.15\% in FTmod, alongside a 14.24\% improvement in PD across the datasets. Moreover, it delivers a layout with zero whitespace, significantly enhancing chip utilization. Figure \ref{fig:total_show} compares Flora’s performance to Corblivar on the n100 dataset, where modules are highlighted in light yellow, and internal macros and standard cell clusters are marked in purple and red, respectively.

Our RT metric is the highest among the four algorithms. This arises because Flora extends traditional approaches by incorporating aspects of detailed placement and routing into the floorplanning stage, increasing runtime. Specifically, we perform rapid component layout within modules and introduce feedthrough metrics to enhance optimization.

\subsection{Ablation Study Results}
To further demonstrate the contributions of each stage of Flora and its utility as a post-processing tool, we conduct an ablation study using Flora to refine Corblivar’s outputs. The results, detailed in Table~\ref{tab:ablation}, outline the optimization achieved at each stage, offering a comprehensive view of our algorithm’s effectiveness.

In the first stage, Flora conducts coarse-grained optimization of HPWL and feedthrough metrics, yielding the best HPWL solution across all three stages. Feedthrough metrics also exhibit a substantial reduction compared to Corblivar’s baseline. While later stages shift focus away from HPWL optimization—resulting in a slight increase—the metrics remain markedly improved over the initial solution.

In the second stage, Flora refines feedthrough metrics by reallocating remaining whitespace, ultimately achieving a zero-whitespace layout. Here, FTmod reaches its optimal value across all datasets, with nearly half also attaining the best FTpin scores. Additionally, this stage’s runtime is notably brief, offering a significant improvement over TOFU, which similarly targets whitespace elimination.

In the third stage, Flora optimizes component placement, achieving the highest PD metrics. Furthermore, over half of the datasets record optimal FTpin values at this stage. We attribute this to the correlation between improved component placement and reduced FTpin, underscoring the value of multi-stage joint optimization.

\section{Conclusion}\label{sec:conclusion}
In this paper, we present \textbf{Flora}, a three-stage rectilinear floorplanning framework that integrates component placement and feedthrough metrics to pre-optimize subsequent physical design stages. This approach ensures alignment with practical EDA requirements. Experimental results confirm that Flora outperforms SOTA methods in overall performance, providing a novel foundation for future research into multi-stage joint optimization.

\bibliographystyle{ieeetr}
\bibliography{ref}

\end{document}